\documentclass[a4paper,fleqn,review,12pt]{cas-sc}

\usepackage[numbers,sort&compress]{natbib}

\usepackage{diagbox}

\usepackage{float}
\restylefloat{figure}

\graphicspath{{../}}

\hypersetup{
    colorlinks=true,
    linkcolor=blue,
    urlcolor=blue,
    citecolor=blue
}

\renewcommand{\printorcid}{}

\begin{document}
\let\WriteBookmarks\relax
\def\floatpagepagefraction{1}
\def\textpagefraction{.001}

\shorttitle{RegNC-Mamba3}
\shortauthors{Z. Li et al.}

\title[mode=title,size=\Large,align=\centering]{\texorpdfstring{RegNC-Mamba3: Shared Structural Feature Learning\\Using Non-Causal State Space Principles\\for Multi-Modal Image Registration}{RegNC-Mamba3: Shared Structural Feature Learning Using Non-Causal State Space Principles for Multi-Modal Image Registration}}

\author[1]{Zhikang Li}
\ead{22021110243@stu.xidian.edu.cn}
\credit{Methodology, Validation, Writing -- original draft, Writing -- review and editing}

\author[1]{Yan Wu}
\cormark[1]
\ead{ywu@mail.xidian.edu.cn}
\credit{Writing -- review and editing, Project administration, Supervision, Funding acquisition}

\author[1]{Xin Hu}
\ead{xinhu@stu.xidian.edu.cn}
\credit{Data curation, Investigation}

\author[1]{Yi Dai}
\ead{850488940@qq.com}
\credit{Investigation}

\author[2]{Ming Li}
\ead{liming@xidian.edu.cn}
\credit{Project administration}

\affiliation[1]{organization={Remote Sensing Image Processing and Fusion Group, School of Electronic Engineering, Xidian University},
    city={Xi'an},
    postcode={710071},
    country={China}}

\affiliation[2]{organization={National Key Laboratory of Radar Signal Processing, Xidian University},
    city={Xi'an},
    postcode={710071},
    country={China}}

\cortext[1]{Corresponding author.}

\begin{abstract}
The main challenge of multi-modal image registration is shared structural feature extraction with higher accuracy and lower computational complexity. Compared with a State Space Model (SSM), the State Space Principles (SSP) formulation of Mamba-3 uses a more accurate discretization with lower truncation error while maintaining linear complexity. However, vanilla SSP is designed for natural language processing (NLP) and is not directly suitable for images. To address this issue, we propose a novel non-causal SSP (NC-SSP) with shared structure awareness and develop a coarse-to-fine registration method named RegNC-Mamba3. First, NC-SSP is designed to eliminate the dependence of the integral computation on the scanning path. By combining relative-weight normalization with aggregation over all circular scan paths, NC-SSP constructs a globally hidden state without long-range decay, which provides a wider receptive field and higher efficiency than the original Mamba-3. Second, NC-SSP is modified to strengthen the extraction of cross-modal structural features. When computing hidden states and feature interactions, we construct complex-valued sequences in NC-SSP by applying learnable Log-Gabor kernels, with different orientations to each head for frequency-domain modulation. This improves the perception of shared structural information across modalities and increases both registration accuracy and quality. Experiments on VIS-SAR and VIS-IR demonstrate that RegNC-Mamba3 outperforms the compared detector-based and semi-dense multi-modal registration methods.
\end{abstract}

\begin{keywords}
Multi-modal image registration \sep Structural feature learning \sep Non-causal SSP \sep Mamba-3
\end{keywords}

\maketitle

\section{Introduction}

Image alignment is a prerequisite for many computer vision tasks. Multi-modal imagery has become increasingly important because different sensors provide complementary information. Registration supports image fusion \cite{ref1,ref2,ref3}, image classification \cite{ref4}, and change detection \cite{ref5,ref6}. The problem is nevertheless difficult: image pairs may differ by rotation, translation, or scale, while modality gaps and sensor noise severely interfere with the extraction of shared features from the two images. Synthetic aperture radar (SAR) offers all-weather and multi-temporal imaging but contains multiplicative speckle. Long-wave infrared (IR) sensors operate under all lighting conditions, yet their images often have soft edges and additive noise.

Depending on whether a detector is employed, multi-modal registration methods can be divided into sparse matching and semi-dense or dense matching approaches. Sparse matching generally follows a detect-then-describe pipeline or a joint detection-and-description framework. In the first category, traditional handcrafted methods detect keypoints and compute descriptors from gradient or phase information \cite{ref7,ref8,ref9,ref10,ref11}, whereas patch-based deep learning methods retain the same workflow but use convolutional neural networks (CNNs) for descriptor extraction \cite{ref12,ref13}. The detector itself can also be implemented using a deep learning-based approach \cite{ref14,ref15}. Joint detection-and-description networks are often built on a Siamese CNN encoder and predict keypoint locations and descriptors simultaneously \cite{ref16,ref17}. Multi-modal registration nevertheless requires shared structural features, whereas CNN-based sparse representations operate on limited receptive field support and are not explicitly aligned across modalities. Graph matching methods introduce global aggregation and cross-image interaction through graph neural networks or Transformers \cite{ref18}, but their results still depend on the input keypoints and descriptors.

Detector-free methods skip feature detection. Most Transformer-based semi-dense matchers \cite{ref19,ref20,ref21} first extract local CNN features, perform coarse interaction and matching at $1/a$ resolution, and then refine correspondences at $1/b$ resolution, typically with $a=8$ and $b=2$. Semi-dense matching methods have also been used for multi-modal image stitching \cite{ref22}. Dense matching follows a similar coarse-to-fine process but predicts all correspondences between the image pair \cite{ref23}, with coarse and fine resolutions of 1/4 and 1. Since every feature participates in interaction and matching, detector-free methods can recover correspondences in weakly textured regions where keypoint detectors often fail. After initial matching, outliers are removed by robust estimators \cite{ref24,ref25,ref26}. Their registration quality is therefore still determined largely by whether the coarse representation captures structures shared by the two modalities.

Mamba is built on selective State Space Models (SSMs) \cite{ref27} and has emerged as a sequence modeling approach with a global receptive field and linear computational complexity. Through its selective scanning mechanism, it has been widely adopted in vision tasks \cite{ref28,ref29,ref30}. The forget gate and state-space block design enable Mamba to outperform linear attention in several sequence-modeling settings \cite{ref31}. A remaining difficulty is that SSMs often train less efficiently than Transformers because they lack a computationally efficient matrix multiplication form. Dao and Gu \cite{ref32} introduced Structured State Space Duality (SSD), which connects structured SSMs with attention variants. Building on SSD, Mamba-2 supports efficient matrix-based computation and improves both performance and efficiency. VSSD \cite{ref33} subsequently adapted SSD to vision tasks by removing the causal mask. Mamba-3 \cite{ref34} replaces the first-order Euler rule with an exponential-trapezoidal discretization using a data-dependent convex combination of both endpoints, and adds complex-valued states via a rotary-embedding equivalence and a low-rank MIMO state; VNCT \cite{ref35} carries this formulation to vision by aggregating the same recurrence non-causally, with low-rank MIMO and 2-D rotary positions. We discuss the details of the above SSMs in Section~\ref{Section 2.2}.

Multi-modal image registration is computationally intensive and remains difficult because of large differences between modalities. Current matching methods generally rely on large-scale training data, so improving training and inference efficiency without sacrificing representational quality is an important problem. Compared with Transformers, SSP keeps the same linear complexity as SSM and SSD while using a more accurate integral approximation, which reduces the global error of the discretized state trajectory. Motivated by these properties, we introduce SSP into a semi-dense matching architecture and propose RegNC-Mamba3. Vanilla SSP is intended for causal language sequences and is therefore not directly suitable for image tokens. We therefore propose NC-SSP for non-causal image tasks. We first eliminate the cumulative multiplication of $A$ in the first-order state equation while retaining the relative numerical weighting. Then we re-derive the integral expression using the trapezoidal rule and aggregation over all circular scan paths. This process removes dependence on the order of adjacent tokens in SSP, and enables each token to interact with information from the full image through a shared hidden state without long-range decay. To make hidden states and cross-modal interaction focus on structural features shared across modalities, we further modulate their frequency representations with learnable multi-orientation Log-Gabor filters following the filtering principle in phase congruency. This construction suppresses modality differences and speckle noise during matching.

The main contributions of this work are as follows:
\begin{itemize} 

\item We propose a novel NC-SSP by removing the path dependence of the SSP integral. Relative-weight normalization and aggregation over all circular scan paths produce a shared, non-decaying hidden state with a global receptive field.

\item We incorporate structural feature awareness into NC-SSP. Learnable multi-orientation Log-Gabor filters are designed to generate complex-valued sequences for each head in hidden state computation and feature interaction.

\item We construct RegNC-Mamba3, a coarse-to-fine detector-free network that integrates cross-modal interaction and multi-scale fusion with NC-SSP. Experiments on VIS-SAR and VIS-IR evaluate the accuracy, robustness, and computational cost of the proposed network.

\end{itemize}

\section{Related Work}
\label{Section II}
\subsection{Multi-modal Image Registration Methods}

A detector-based registration pipeline consists of keypoint detection, descriptor extraction, feature matching, and transform estimation. RIFT \cite{ref10} detects keypoints with FAST \cite{ref36} on a maximum-moment map and describes the resulting points using the six-orientation Maximum Index Map (MIM). Its detector and descriptor are both derived from Log-Gabor filtering and phase congruency, which makes the representation less sensitive to radiometric change and rotation. SDNet \cite{ref37} applies self-distillation so that deep descriptors guide the learning of shallower patch representations. ReDFeat \cite{ref17} follows the joint detection-and-description paradigm and recouples detection and description through mutual weighting based on the R2D2 framework \cite{ref16}. However, within the detector-based pipeline, keypoint repeatability remains a source of failure. Handcrafted descriptors are less robust to nonlinear radiometric changes and noise, while patch-based methods have difficulty extracting discriminative descriptors when the patches contain little structural information. Joint detection-and-description reduces the mismatch between keypoint detection and description, but it still suffers from noise and modality differences in multi-modal image registration.

Detector-free methods skip keypoint selection and directly match feature maps in a coarse-to-fine process. LoFTR \cite{ref19} introduced Transformer-based semi-dense matching: it establishes initial correspondences on 1/8-resolution features and refines them in local windows on 1/2-resolution. ASpanFormer \cite{ref20} extends this idea with local-global attention and adaptive attention span at the coarse level. XoFTR \cite{ref21} targets VIS-IR matching and fuses features from different scales with window-based cross-attention. For multi-modal data, these methods still require a token representation that is both globally informative and robust to modality differences. The accuracy of detector-free methods is closely tied to coarse matching. If the coarse similarity matrix lacks reliable peaks, the subsequent local windows contain few correct candidates, and a more expressive fine stage cannot recover the missing correspondences. This characteristic makes coarse cross-modal feature interaction the main target of the present work.

\subsection{SSM, SSD, and SSP} \label{Section 2.2}
State space models (SSMs) have received increasing attention in vision because they can aggregate sequence information over a global range while preserving linear complexity. In a vanilla SSM, the first-order linear differential equation is:
\begin{align}\label{Eq.1}
h'(t) & = Ah(t)+BX(t) \nonumber\\
y(t) &= Ch(t)
\end{align}
To apply this model to a discrete sequence, the continuous equation is commonly discretized with Euler's rule:
\begin{align}\label{Eq.2}
h_t & = \bar{A} h_{t-1}+\bar{B} X_t \nonumber\\
y_t & = Ch_t
\end{align}

For image tasks, Zhu et al.~\cite{ref38} identified twelve global scanning directions for image SSMs under a specified definition, and Mamba-based methods usually combine a subset of these directions to enlarge the receptive field. SSMs have also been applied to semi-dense image matching for cross-image interaction. JamMa \cite{ref39} concatenates two images horizontally and vertically and scans the joint sequence in four directions. Because the two images are placed in one sequence, all four scans follow handcrafted scan paths and may mix content from the other image, disrupting the global representation of each image.

To implement the SSM scanning process more efficiently, the Structured State Space Duality (SSD) \cite{ref32} formalizes the relation between SSMs and attention. An SSD layer in Mamba-2 can be interpreted as masked linear attention:
\begin{align}\label{Eq.3}
M &= L\circ (CB^{\top})\nonumber\\
Y &= MX
\end{align}

Structured matrix algorithms make SSD training and inference faster with a global receptive field. Like SSM, SSD was developed for causal sequences such as those in language modeling, whereas image tokens have no natural ordering. VSSD \cite{ref33} addresses this mismatch by converting SSD to a non-causal form. It removes the cumulative multiplication of $A$, combines bidirectional scanning paths, and shares hidden states across all tokens so that no lower-triangular mask remains. Its aggregation can be written as:
\begin{align}\label{Eq.5}
H &= \sum_{j=1}^{i} \frac{1}{A_j} B_jx(j) +\sum_{j=-L}^{-i} \frac{1}{A_{-j}} B_{-j}x(-j)  \nonumber \\
& = \sum_{j=1}^{L} \frac{1}{A_j} B_jx(j) \nonumber  \\
Y &= CH 
\end{align}

SSM and SSD nevertheless use Euler's rule, which approximates the integral over each step by a rectangle and therefore accumulates first-order truncation error over long sequences. SSP \cite{ref34} replaces this approximation with generalized trapezoidal discretization, reducing the numerical error between the continuous integral and its discrete approximation. SSP also supports complex-valued hidden states, which can represent rotational state transitions, and adopts a multi-input multi-output formulation to improve inference efficiency. The hidden-state update of Mamba-3 with generalized trapezoidal discretization is:
\begin{align}\label{Eq.4}
\tilde{h} _t &= e^{\bigtriangleup _{t} A _{t}}\tilde{h} _{t-1} + (1-\lambda _t)\bigtriangleup _{t}e^{\bigtriangleup _{t} A _{t}}
\bar{B} _{t-1}x_{t-1} + \lambda _t\bigtriangleup _{t}\bar{B} _{t}x_{t} \nonumber\\
&=\alpha_t \tilde{h} _{t-1}+ \beta _t\bar{B} _{t-1}x_{t-1} + \gamma_t\bar{B} _tx_t \nonumber\\
y_t & = \bar{C} _{t}^{\top } \tilde{h} _t
\end{align}

Similar in motivation to our recent work, VNCT \cite{ref35} proposes NC-M3 for image tasks. NC-M3 retains the trapezoidal recurrence of Mamba-3 and attempts to remove its causal constraint. The process of NC-M3 can be calculated as:
\begin{align}
w_j &=\mathrm{softmax}_L(\gamma_j/\sqrt{N})+\mathrm{softmax}_L(\beta^{+}_j/\sqrt{N})\nonumber\\
G &= \sum_{j=1}^{L} w_j \tilde{x}_j\tilde{B} _j\nonumber\\
y &= CG
\end{align}
where $\beta^{+}_j = \beta_{j+1}$. However, $\beta_{j+1}$ cannot be known if the sequence is order-invariant, as the next item is not identifiable without a fixed scan path. Moreover, the boundary value $\beta_{L+1}$ is not defined. Therefore, the permutation-invariant aggregation and isotropic ERF arguments proposed in VNCT lack their stated basis.

To ensure full permutation invariance over the token sequence of Mamba-3, we do not adopt the formulation of VNCT. Drawing inspiration from VSSD, which preserves relative weights while discarding absolute magnitudes, as well as the trapezoidal rule in Mamba-3, we propose our novel NC-SSP. Relative-weight normalization first eliminates multiplicative long-range decay. The sequence is then connected into a directed cycle, and the update is averaged over all circular scan paths when computing hidden states. As a result, the shared state does not depend on any fixed scan direction, and each token can access global context in a single interaction. This property is important for registration because the global receptive field of a point must remain invariant to changes in the scan direction.

The second component of NC-SSP is structural feature awareness. We modulate the frequency-domain representations with learnable multi-orientation Log-Gabor filters when computing hidden states and cross-modal interaction, following the multi-orientation filtering used in phase congruency. Phase congruency responds to local structural features rather than absolute pixel intensity values, which is beneficial when optical, SAR, and IR images have different radiometric responses. In NC-SSP, this modulation also gives the rotation angle of the complex-valued state an interpretable relation to image orientation. The complete derivation and filter construction are presented in Sections \ref{Section 3.1} and \ref{Section 3.2}.

\section{Methodology}
This section first derives NC-SSP, including its relative-weight normalization and aggregation over all circular scan paths. It then describes the frequency-domain construction of structural-feature-aware $B$ and $C$ in the hidden state, and finally presents the complete RegNC-Mamba3 architecture.

\subsection{Derivation of NC-SSP}\label{Section 3.1}

Firstly, we normalize the state transition in the continuous SSM so that only the relative weighting of $A$ and $1$ is retained. The resulting first-order equation is:

\begin{align}\label{Eq.6}
\frac{\mathrm{d} h(t)}{\mathrm{d} t}=h(t) + \frac{B(t)X(t)}{A(t)}  
\end{align}

Integrating Eq.~\eqref{Eq.6} over $[t_0,t_1]$ gives:
\begin{align}\label{Eq.7}
h(t_{1})-h (t_{0}) & = \int_{t_{0}}^{t_{1}} h(t)dt+\int_{t_{0}}^{t_{1}} \frac{B(t)X(t)}{A(t)}dt 
\end{align}

\begin{figure}[H]
	\centering
	\includegraphics[width=0.6\textwidth]{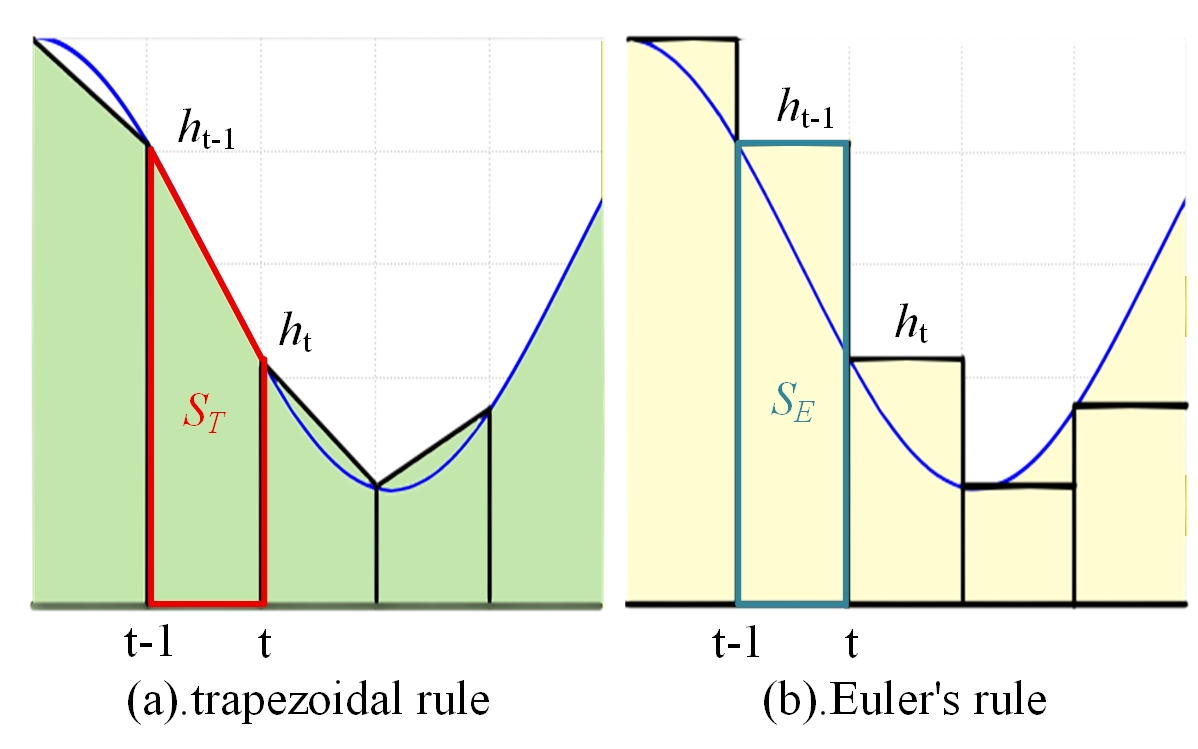}
	\caption{Schematic diagram of integral approximation: (a) generalized trapezoidal discretization in SSP; (b) Euler's rule in SSMs.}
	\label{fig.1}
\end{figure}

As in Mamba-3, the integral is approximated by generalized trapezoidal discretization, as illustrated in Fig.~\ref{fig.1}:
\begin{align}\label{Eq.8}
h_{t}-h_{t-1}=&[\lambda_{t}h_{t}+(1-\lambda_{t})h_{t-1}]\Delta  t \nonumber\\
+&[\lambda_{t}\frac{B_{t}X_{t}}{A_{t}} +(1-\lambda_{t})\frac{B_{t-1}X_{t-1}}{A_{t-1}}]\Delta  t
\end{align}

Let $Z_{t}=B_{t}X_{t}/A_{t}$, $m_{t}=(1-\lambda_{t})\Delta t$, $n_{t}=\lambda_{t}\Delta t$, and $o_{t}=1-\lambda_{t}\Delta t$. Eq.~\eqref{Eq.8} can then be rearranged as:
\begin{align}\label{Eq.9}
h_{t}=&\frac{1+m_{t} }{o_{t} } h_{t-1}+\frac{n_{t}}{o_{t}} Z_{t}+\frac{m_{t}}{o_{t}} Z_{t-1}
\end{align}

Eq.~\eqref{Eq.9} contains a multiplicative coefficient on $h_{t-1}$, which would cause long-range decay. We therefore apply relative-weight normalization again, following VSSD:
\begin{align}\label{Eq.10}
h_{t} & = h_{t-1} + \frac{n_t}{(1+m_t)}Z_{t} +\frac{m_t}{(1+m_t)}Z_{t-1} \nonumber\\
& = h_{t-1} + \beta_{t}Z_{t} + \gamma _{t}Z_{t-1}
\end{align}
where $\beta_{t}=n_{t}/(1+m_t)$ and $\gamma _{t}={m_{t}}/(1+m_t)$.

To remove the endpoints of the sequence, we connect the first and last elements into a directed cycle. The hidden state for one complete traversal is then:
\begin{align}\label{Eq.11}
H=& (\beta _{1}Z_{1} +\gamma _{1} Z_{N}) + \sum_{i=1}^{N-1} (\beta _{i+1} Z_{i+1}+ \gamma _{i+1}Z_{i})
\end{align} 

In Eq.~\eqref{Eq.11}, the term $\gamma_iZ_{i-1}$ still depends on the chosen traversal order. Selecting several fixed scan paths is less effective in image registration when the two images have different transformations, since the paths are not spatially invariant and increase the computational overhead. Instead, we aggregate all circular traversals. For a sequence of $N$ elements, cyclic shifts are treated as equivalent, yielding $(N-1)!$ distinct circular permutations. Each permutation is linearized into a scan path and included in the following average:
\begin{align}\label{Eq.12}
H_{NC} & = \frac{1}{(N-1)!} \sum_{j= 1}^{(N-1)!} \operatorname{ScanPath}_{j}(H_{j})\nonumber  \\
& = \frac{1}{(N-1)!} \sum_{j= 1}^{(N-1)!} \operatorname{ScanPath}_{j}[\sum_{i = 1}^{N} (\beta _{i}Z_{i}+\gamma_{i}Z_{i-1})]\nonumber\\ 
& = \sum_{i = 1}^{N} \beta _{i}Z_{i} + \frac{(\sum_{i = 1}^{N}\gamma_{i}) {\textstyle \sum_{i = 1}^{N}Z_{i}}-\sum_{i = 1}^{N}\gamma_{i}Z_{i}}{N-1}\nonumber\\
& = H_{p} + H_{n} 
\end{align}

The resulting hidden state $H_{NC}$ is shared by all tokens and averaged over every possible traversal order, so it is independent of the scanning direction. The same aggregate can be computed in one matrix multiplication, allowing Eq.~\eqref{Eq.12} to be rewritten as follows:
\begin{align}\label{Eq.13}
H_{NC} &= B^{\top } \Lambda X \\
\Lambda &= \operatorname{diag}[\frac{(N-1)\beta_j  +  {\textstyle (\sum_{i=1}^{N}\gamma _{i})- \gamma}_j}{(N-1)A_j}]_{j=1}^{N} 
\end{align}

The complete NC-SSP output is:
\begin{align}\label{Eq.15}
Y =& CH_{NC} + DX
\end{align}

\subsection{Shared Structural Feature-Aware NC-SSP Block}\label{Section 3.2}
\begin{figure}[H]
	\centering
	\includegraphics[width=0.9\textwidth]{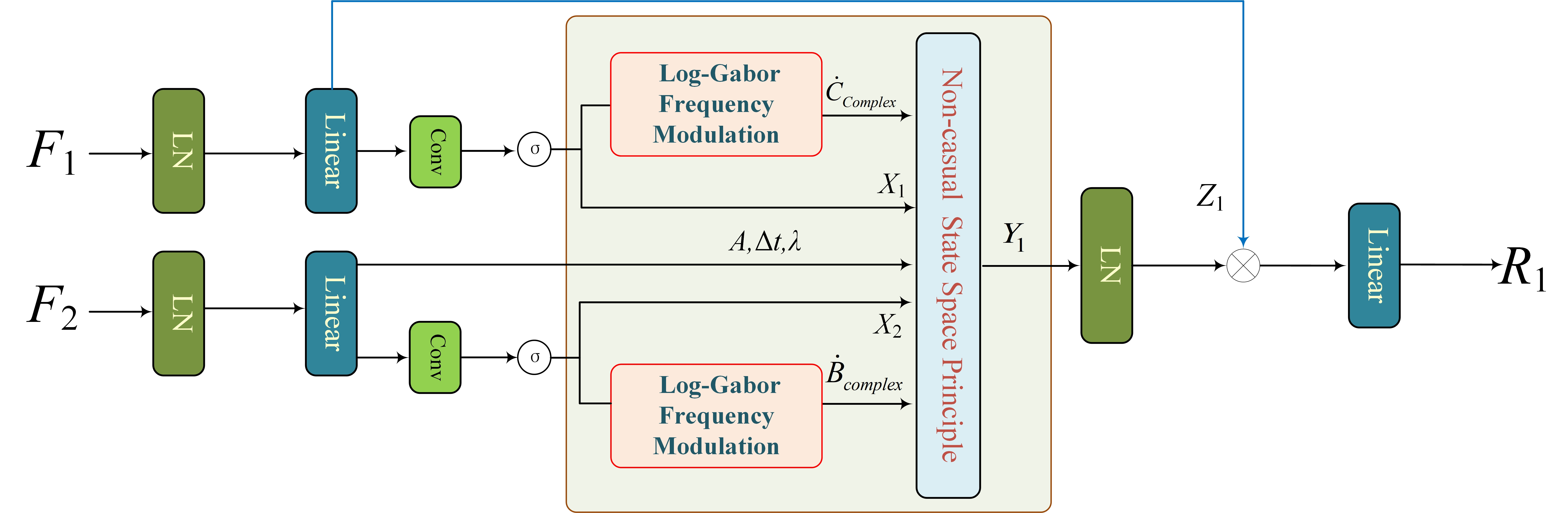}
\caption{Architecture of the shared structural-feature-aware NC-SSP layer.}
	\label{fig.2}
\end{figure}
 
To retain the SSP computation while making its token mixer sensitive to shared cross-modal structural features, we adapt the multi-orientation Log-Gabor modulation used in phase congruency (PC) \cite{ref10,ref40}, whose filters are learnable. The complete block is shown in Fig.~\ref{fig.2}. We first obtain three real-valued sequences from linear projections:
\begin{align}\label{Eq.16}
B=&\operatorname{Linear}(x_{2}),X=\operatorname{Linear}(x_{2}), \nonumber \\
C=&\operatorname{Linear}(x_{1})
\end{align}

As with self-attention and cross-attention, NC-SSP is called intra-modal when $x_1$ and $x_2$ come from the same image, and cross-modal when they come from images of different modalities.

We use $B$ to describe the construction. Its channels are split into $N$ heads, and each head has an orientation angle $\theta_n$ initialized from one of $N$ predefined directions. The functional forms of the Log-Gabor and low-pass responses are retained, while their cutoff and bandwidth parameters are learned. The filtered sequence is:
\begin{align}\label{Eq.17}
W^{n} &= W_{Log-Gabor}^{n} \cdot W_{lowpass}^{n}, \nonumber\\
\dot{B} _{freq} & = \operatorname{FFT}(B), \nonumber\\
\dot{B} _{filt}^{n} & = \operatorname{iFFT}(W^{n}F(\theta_{n})\dot{B} _{freq})
\end{align}

The kernels $W_{Log-Gabor}^{n}$, $W_{lowpass}^{n}$, and $F(\theta_n)$ follow the standard multi-orientation PC formulation \cite{ref10,ref40}; only their key filter parameters are learnable. The real and imaginary components are projected and activated as:
\begin{align}\label{Eq.18}
B_{I}^{n} & = \phi [\operatorname{proj}(\operatorname{Imag}(\dot{B}_{filt}^{n}))+bias], \nonumber\\
B_{R}^{n} & = \phi [\operatorname{proj}(\operatorname{Real}(\dot{B}_{filt}^{n}))+bias],
\end{align}
where $\mathrm{bias}=\varepsilon B$ serves as a spatial residual bias, $\phi$ is a kernel function, and $\mathrm{proj}$ is a linear projection. The complex-valued representation of head $n$ is:
\begin{align}\label{Eq.19}
\dot{B} _{complex}^{n} & = B_{R}^{n}+jB_{I}^{n}   
\end{align}

Applying the same construction to $C$ gives:
\begin{align}\label{Eq.20}
\dot{C} _{complex}^{n} & = C_{R}^{n}+jC_{I}^{n}   
\end{align}

The complex hidden state is obtained by applying the NC-SSP aggregation in Eq.~\eqref{Eq.13} separately to the real and imaginary components:
\begin{align}\label{Eq.21}
\dot{H} _{complex}^{n} & = (B_{R}^{n})^{\top} \Lambda X+j(B_{I}^{n})^{\top} \Lambda X, \nonumber\\
H _{R}^{n} &= (B_{R}^{n})^{\top} \Lambda X,\nonumber\\
H _{I}^{n} &= (B_{I}^{n})^{\top} \Lambda X
\end{align}

Taking the real part of $\dot{C}_{complex}^{n}\dot{H}_{complex}^{n}$ gives:
\begin{align}\label{Eq.22}
Y_{M}^{n} &=\operatorname{Real}(\dot{C} _{complex}^{n}\cdot \dot{H} _{complex}^{n})\nonumber \\
& = C_{R}^{n} H_{R}^{n} - C_{I}^{n} H_{I}^{n}
\end{align}

The head outputs are concatenated, and a residual branch is incorporated as in Eq.~\eqref{Eq.15}:
\begin{align}\label{Eq.23}
Y=Y_{M} +DX
\end{align}

\begin{figure}[H]
	\centering
	\includegraphics[width=0.6\textwidth]{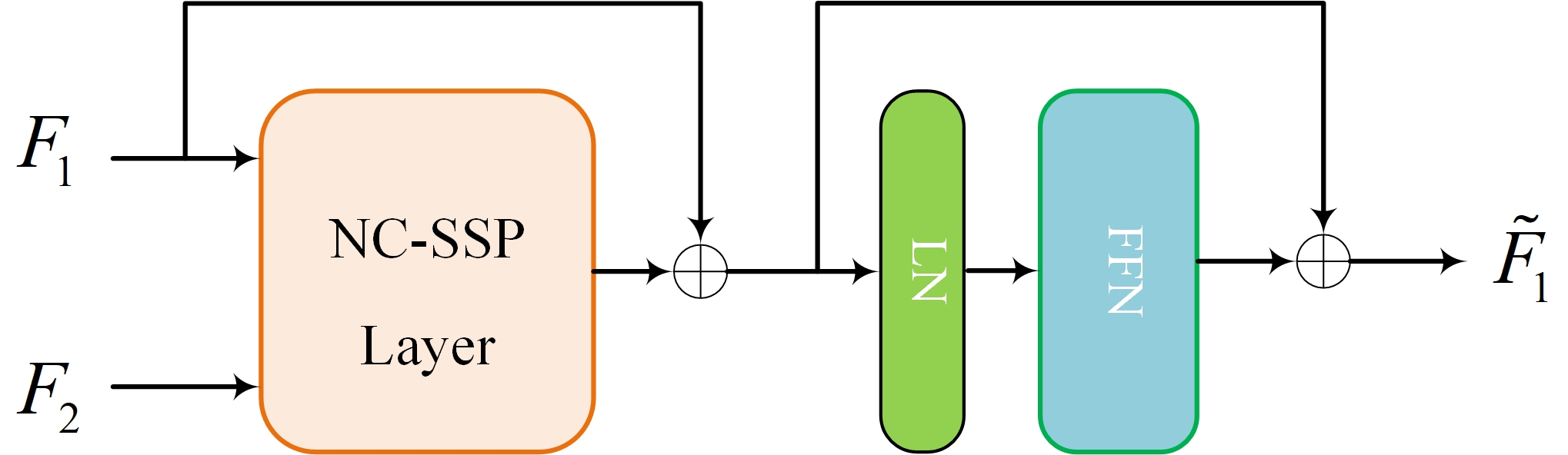}
	\caption{The architecture of our NC-SSP block.}
	\label{fig.3}
\end{figure}

Each NC-SSP block follows the Transformer block layout, with an NC-SSP token mixer followed by a feed-forward network, which is shown in Fig.~\ref{fig.3}.

\subsection{Architecture of RegNC-Mamba3}\label{Section 3.3}
\begin{figure}[H]
	\centering
	\includegraphics[width=\textwidth]{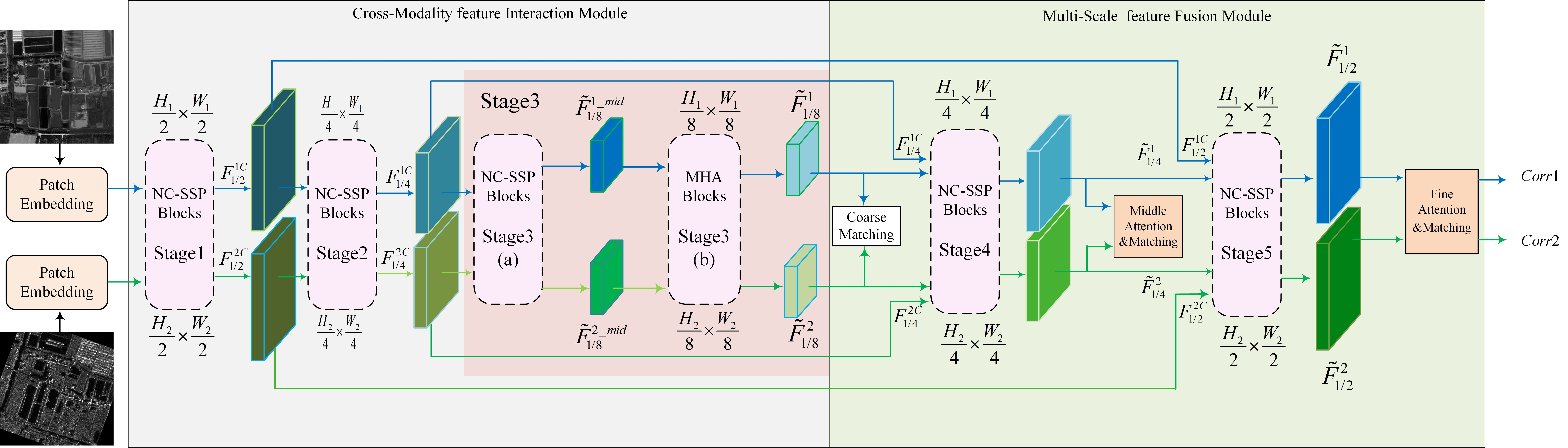}
\caption{Architecture of RegNC-Mamba3. The Cross-Modal Feature Interaction (CMI) module extracts shared structural features at scales 1/2, 1/4, and 1/8 in Stages~1--3. Stage~3(a) contains NC-SSP blocks and Stage~3(b) contains MHA blocks. The Multi-Scale Feature Fusion (MSF) module fuses features at scales $[1/8, 1/4]$ in Stage~4 and $[1/4, 1/2]$ in Stage~5. Coarse correspondences are computed at 1/8 resolution; $5\times5$ and $7\times7$ window attention is then used at 1/4 and 1/2 resolution to obtain intermediate and fine correspondences.}
	\label{fig.4}
\end{figure}

Figure~\ref{fig.4} shows the overall architecture. A Cross-Modal Feature Interaction (CMI) module operates at scales 1/2, 1/4, and 1/8, and a Multi-Scale Feature Fusion (MSF) module progressively combines features from 1/8 to 1/2 resolution. Coarse correspondences are estimated at 1/8 resolution. Window attention with sizes $5\times5$ and $7\times7$ is then applied at 1/4 and 1/2 resolution to obtain intermediate and fine correspondences.

\subsubsection{Cross-Modal Feature Interaction (CMI) Module}
The CMI module comprises Stages~1--3. Each stage contains one intra-modal and one cross-modal NC-SSP operation per input:
\begin{align}\label{Eq.24}
x_{s}^{1\_self} &= \operatorname{NC-SSP}(x_{s}^{1} ,x_{s}^{1}),\nonumber\\
x_{s}^{2\_self} &= \operatorname{NC-SSP}(x_{s}^{2} ,x_{s}^{2}),\nonumber\\
x_{s}^{1C} &= \operatorname{NC-SSP}(x_{s}^{1\_self} ,x_{s}^{2\_self}),\nonumber\\
x_{s}^{2C} &= \operatorname{NC-SSP}(x_{s}^{2\_self} ,x_{s}^{1\_self})
\end{align}
where the scales $s\in [1/2,1/4,1/8]$ correspond to Stages~1, 2, and 3(a).

SSP can be interpreted as a form of linear attention and is efficient for global aggregation, but its refinement capability is weaker than that of multi-head attention. We therefore use multi-head attention (MHA) in Stage~3(b) to produce more discriminative features for coarse matching.
\subsubsection{Multi-Scale Feature Fusion (MSF) Module}
The MSF module progressively fuses the multi-scale CMI features and produces the final 1/2-resolution representation. It contains Stages~4 and~5, both built from NC-SSP blocks. Stage~4 computes:
\begin{align}\label{Eq.25}
\tilde{F}_{1/4}^{1} &= \operatorname{NC-SSP}(F_{1/4}^{1C},\tilde{F}_{1/8}^{1}),\nonumber\\
\tilde{F}_{1/4}^{2} &= \operatorname{NC-SSP}(F_{1/4}^{2C},\tilde{F}_{1/8}^{2}) 
\end{align}
where $\tilde{F}_{1/8}^{1}$ and $\tilde{F}_{1/8}^{2}$ are coarse features from Stage~3(b), and $F_{1/4}^{1C}$ and $F_{1/4}^{2C}$ are the cross-modal features from Stage~2.

Stage~5 performs the fusion from 1/4 to 1/2 resolution:
\begin{align}\label{Eq.26}
\tilde{F}_{1/2}^{1} &= \operatorname{NC-SSP}(F_{1/2}^{1C},\tilde{F}_{1/4}^{1}),\nonumber\\
\tilde{F}_{1/2}^{2} &= \operatorname{NC-SSP}(F_{1/2}^{2C},\tilde{F}_{1/4}^{2}) 
\end{align}
where $\tilde{F}_{1/4}^{1}$ and $\tilde{F}_{1/4}^{2}$ are the Stage~4 outputs, and $F_{1/2}^{1C}$ and $F_{1/2}^{2C}$ are the cross-modal features from Stage~1.

The Stage~5 output therefore combines information from all five stages at 1/2 resolution and is used in the final refinement.

\subsubsection{Coarse-to-Fine Refinement}
Following the refinement procedure of \cite{ref21}, coarse matching indices $idx_1$ and $idx_2$ are obtained by applying softmax and argmax to $sim_{1/8}^{B}$. The corresponding $5\times5$ windows are then extracted from the Stage~4 features $[\tilde{F}_{1/4}^{1},\tilde{F}_{1/4}^{2}]$. For a window pair of a correspondence, let $[f_{5\times5}^{1M},f_{5\times5}^{2M}]$ denote the intermediate features after Stage~4 window attention. Their normalized similarity is:
\begin{align}\label{Eq.27}
sim_{5\times 5}^{Mid} & = \operatorname{softmax}[\frac{(f_{5\times 5}^{1M}) \cdot (f_{5\times 5}^{2M})^{\top }}{\tau }]
\end{align}

The displacement $[\Delta mid_1,\Delta mid_2]$ is selected by $\operatorname{argmax}(sim_{5\times5}^{Mid})$, and the center of the fine-level window is updated as:
\begin{align}\label{Eq.28}
mid_1 &= \mathrm{coarseIdx}_1 + \bigtriangleup mid_1,\nonumber \\
mid_2 &= \mathrm{coarseIdx}_2 + \bigtriangleup mid_2
\end{align}

The corresponding $7\times7$ windows are then extracted from the 1/2-resolution features. Let $[f_{7\times7}^{1F},f_{7\times7}^{2F}]$ be the features after Stage~5 window-attention, and let:
\begin{align}\label{Eq.29}
sim_{7\times 7}^{Fine} & = \operatorname{softmax}[\frac{(f_{7\times 7}^{1F}) \cdot (f_{7\times 7}^{2F})^{\top }}{\tau }]
\end{align}

Fine correspondences are obtained by $\operatorname{argmax}(sim_{7\times7}^{Fine})$. Training uses supervised cross-entropy loss at the coarse, intermediate, and fine stages:
\begin{align}\label{Eq.30}
L_{total}=& L_{Coarse} +  L_{mid} + L_{Fine}\nonumber\\
 = &CE(sim_{1/8}^{B},\mathrm{GT}_{\mathrm{coarse}}) + CE(sim_{5\times 5}^{Mid},\mathrm{GT}_{\mathrm{mid}})\\
 + &CE(sim_{7\times 7}^{Fine},\mathrm{GT}_{\mathrm{fine}})\nonumber
\end{align}

\section{Experiments}
\subsection{Experimental Setup}
\subsubsection{Datasets}

We evaluate RegNC-Mamba3 on two multi-modal datasets: VIS-SAR and VIS-IR.

\textbf{VIS-SAR.} The OSDataset \cite{ref41} contains co-registered SAR and optical image pairs from GaoFen-3 and Google Earth. The images have a 1-m ground sampling distance and a spatial size of $512\times512$ pixels. We use 2,011 pairs for training and 424 pairs for testing.

\textbf{VIS-IR.} DroneVehicle \cite{ref42} provides aligned IR and VIS images of street scenes under daytime and night-time conditions, each of size $640\times512$ pixels. We use 4,141 pairs for training and 1,696 pairs for testing.

\subsubsection{Evaluation Metrics}
Registration performance is measured by root mean square error (RMSE), number of correct matches (NCM), and successful matching rate (SMR).

\textit{RMSE:} RMSE measures the average displacement error of the estimated homography:
\begin{align} \label{Eq 31}
RMSE = \sqrt{\frac{1}{N}\sum_{i=1}^{N}(H_{e}M^{i}(x,y)-H_{gt}M^{i}(x,y))^{2} }
\end{align}
where $H_e$ is the homography estimated by MAGSAC \cite{ref25} with a reprojection threshold of 2 pixels and $10^6$ iterations, $H_{gt}$ is the ground-truth homography, and $\mathcal{M}$ is a grid of $N=961$ points sampled with spacing $(H/32,W/32)$.

\textit{NCM:} NCM is the number of correspondences whose Euclidean distance to the ground-truth location is less than 2 pixels.

\textit{SMR:} SMR is the percentage of test pairs registered with RMSE below 5 pixels.

Average RMSE (aRMSE) and average NCM (aNCM) are computed over successfully registered pairs only, whereas all test pairs remain in the denominator of SMR.

\subsubsection{Network and Training Setup}
The network has five stages with channel dimensions $[64, 96, 256, 96, 64]$. The feature-map resolutions from Stage~1 to Stage~5 are $[1/2, 1/4, 1/8, 1/4, 1/2]$ of the input size. All models are implemented in PyTorch. Training uses AdamW with cosine annealing. The learning rate is set to $1\times10^{-4}$, and the number of training epochs is set to 25. Data augmentation includes random rotation in $[-40^{\circ},40^{\circ}]$ and random scaling in $[0.7, 1.3]$, with a batch size of 1.

\subsubsection{Comparison Methods}

We compare RegNC-Mamba3 with three detector-based and three detector-free methods. The detector-based set covers handcrafted, patch-based learned, and joint detection-and-description features: RIFT \cite{ref10}, SDNet \cite{ref37}, and ReDFeat \cite{ref17}. For RIFT, 5,000 keypoints are detected by FAST on the maximum-moment map and described on $96\times96$ patches of the MIM feature map. SDNet uses the same 5,000 FAST keypoints and $64\times64$ patches for its CNN descriptor. For ReDFeat, $7\times7$ non-maximum suppression is applied to the score maps, and the 5,000 highest-scoring keypoints are retained.

The detector-free set consists of LoFTR \cite{ref19}, XoFTR \cite{ref21}, and JamMa \cite{ref39}. A fixed confidence threshold is not transferable across these methods and datasets, so all semi-dense methods return at most the top $K$ fine-stage correspondences. The default budget is $K=2000$, and sensitivity to $K$ is examined separately in Table~\ref{tab.4}.

\subsection{Ablation Study}
\subsubsection{Backbone and Structural Feature Awareness}
We isolate the effects of non-causal aggregation and structural feature awareness (SFA) by replacing only the token mixer while keeping the remaining network fixed. Table~\ref{tab.1} compares six variants:
\begin{enumerate}
\item VSSD: non-causal SSD with Euler discretization, as used in Mamba-2;
\item VSSD+SFA: VSSD with the same Log-Gabor modulation of complex-valued $B$ and $C$;
\item Ori-SSP: VMamba-style Mamba-3 with four scan directions and long-range decay;
\item Ori-SSP+SFA: the same Mamba-3 backbone with structural feature awareness;
\item NC-SSP: the proposed non-causal SSP without structural feature awareness;
\item NC-SSP+SFA: the complete block used in RegNC-Mamba3.
\end{enumerate}

\begin{table}[pos=H]
\caption{Ablation of the backbone and structural feature awareness on VIS-SAR and VIS-IR.\label{tab.1}}
\small
\centering
\begin{minipage}{1\textwidth}
\centering
\resizebox{\textwidth}{!}{%
\begin{tabular}{|c|c|c|c|c|c|c|c|}
\hline
Dataset & Method & aRMSE$\downarrow$ &	aNCM$\uparrow$ & SMR(\%)$\uparrow$ & FLOPs (G) & time(ms) & memory(MB)\\
\hline
\multirow{6}{*}{VIS-SAR} & VSSD	& 2.657 & 325 & 84.4 & 588.2 & 92.05 & 745 \\
&  VSSD+SFA & \underline{2.266} & \underline{689.2} & \underline{89.4} & 588.3 & 92.53 & 750  \\
&  Ori-SSP & 2.712 & 464.8 & 71.9  & 601.3 & 669.09 & 1724 \\
& Ori-SSP + SFA & 2.442 & 541.1& 89.2  & 601.4 &  687.53 & 1743 \\
&NC-SSP & 2.595 & 258.4 & 82.3  & 627.0 & 93.05 & 761  \\
&NC-SSP+SFA & \textbf{1.901} & \textbf{724.3} & \textbf{97.6} & 627.1 & 93.54 & 765 \\
\hline
\multirow{6}{*}{VIS-IR}  & VSSD	& 1.101 & 1541.8 & 98.5 & 764.4 &115.57 & 884 \\
&  VSSD+SFA & \underline{1.062} & \underline{1590.9} & \textbf{98.7} & 764.5 &115.70 & 889  \\
&  Ori-SSP & 1.156 & 1399.8 & 94.8  & 780.8 &847.81 & 2142 \\
& Ori-SSP + SFA & 1.233 & 1477.7 & \textbf{98.7}  & 780.9 & 869.82 & 2164 \\
&NC-SSP & 1.203 & 1211.5 & 98.3  & 812.9 & 114.99 & 901  \\
&NC-SSP+SFA & \textbf{0.911} &\textbf{ 1625.0 }& \textbf{98.7 }& 813.0 & 129.45 & 906 \\
\hline

\end{tabular}%
}
\end{minipage}
\end{table}

The VMamba-style Mamba-3 computes hidden states along four separate scan paths, which results in many small-kernel operations and high memory consumption. VSSD and NC-SSP instead aggregate all tokens into shared hidden states, so the interaction can be implemented with larger matrix operations. Relative to Ori-SSP, NC-SSP without SFA uses 44.1\% of the GPU memory and runs $7.19\times$ faster on $512\times512$ VIS-SAR inputs. On $640\times512$ VIS-IR inputs, the corresponding memory ratio is 42.1\% and the speedup is $7.37\times$. Adding SFA introduces two hidden-state computations for the real and imaginary components and frequency-domain modulation at each stage. Its FLOP and runtime overhead is small relative to the full network because the subsequent window attention accounts for most of the computation, while SMR and aNCM increase for all three backbones. One exception is that SFA slightly increases the IR aRMSE of Ori-SSP, but aNCM and SMR are clearly improved.

The ablation also reveals an important relationship between NC-SSP and VSSD. NC-SSP alone aggregates the current-token term $H_p$ and the order-averaged neighbor term $H_n$, whereas the VSSD aggregation in Eq.~\eqref{Eq.5} contains only the former. Despite this additional global context, NC-SSP alone sometimes performs slightly worse than VSSD: its SAR SMR is 82.3\% versus 84.4\%, and its IR aRMSE is 1.203 versus 1.101. The advantage appears after structural feature awareness is added. NC-SSP+SFA reaches the lowest aRMSE and highest aNCM on both datasets, the highest SAR SMR, and a tied highest IR SMR. This comparison indicates that path-invariant aggregation and structural feature awareness are complementary.

\subsection{Image Registration Performance}
\subsubsection{Performance on Two Datasets}

\begin{table}[pos=H]
\caption{Average RMSE, NCM, and SMR of RegNC-Mamba3 and six comparison methods on VIS-SAR and VIS-IR.}
\label{tab.2}
\centering
\begin{minipage}{\textwidth}
\centering
\resizebox{\textwidth}{!}{%
\begin{tabular}{|c|c|c|c|c|c|c|c|c|}
\hline
    Dataset & Metrics & RIFT &	SDNet & ReDFeat & LoFTR & XoFTR & JamMa & RegNC-Mamba3\\
\hline
\multirow{3}{*}{VIS-SAR}
& aRMSE$\downarrow$ &3.104 & 2.984 & \underline{2.400} & 2.584 & 3.075 & 3.033 & \textbf{1.901} \\

& aNCM$\uparrow$ & 93.3 & 78.3 & 115.6 & 281.4 &  \underline{292.4} & 93.0 & \textbf{724.3}\\
& SMR(\%)$\uparrow$ & 14.4 & 55.0 & \underline{75.7} & 74.8 & 60.1 & 60.4 & \textbf{97.6}\\
\hline
\multirow{3}{*}{VIS-IR} 
& aRMSE$\downarrow$ & 2.009 & 1.822 & 1.168 & 0.983 & \underline{0.942} & 1.119 & \textbf{0.911}\\
& aNCM$\uparrow$ & 513.5 & 116.0 & 357.1 & 1148.3 & \underline{1222.8} & 732.4 & \textbf{1625.0} \\
& SMR(\%)$\uparrow$ & 66.6 & 75.0 & 93.8 & 94.6 & 96.9 & \underline{98.1} & \textbf{98.7}\\
\hline

\end{tabular}%
}
\end{minipage}
\end{table}

Table~\ref{tab.2} compares RegNC-Mamba3 with six methods on VIS-SAR and VIS-IR. Among the detector-based methods, RIFT has the lowest SMR on VIS-SAR because its handcrafted phase-based detector and descriptor are difficult to handle speckle and weak edges. SDNet improves the IR SMR to 75.0\% but reaches only 55.0\% on SAR. ReDFeat is the strongest sparse method, with aRMSE values of 2.400 and 1.168 and SMR values of 75.7\% and 93.8\%. Its larger error on SAR indicates that a CNN backbone remains difficult to detect enough reliable keypoints under strong radiometric variation and speckle.

For detector-free matching methods, LoFTR and XoFTR use coarse self- and cross-attention and can therefore establish global correspondences. XoFTR improves on LoFTR on VIS-IR, with aRMSE decreasing from 0.983 to 0.942 and SMR increasing from 94.6\% to 96.9\%. On VIS-SAR, XoFTR reaches only 60.1\% SMR, showing that its multi-scale fusion is less effective in the presence of severe SAR noise. JamMa reaches 60.4\% SAR SMR but 98.1\% IR SMR. Its joint-scan representation is therefore reliable when the modality gap is moderate, but its SAR performance follows the limitation discussed in Section~\ref{Section 2.2}. Since all semi-dense methods refine local windows around coarse matches, an unreliable coarse similarity map is difficult to correct at the fine stage.

\begin{table}[pos=H]
\caption{Percentage of successfully registered pairs below 1-, 3-, and 5-pixel RMSE thresholds on VIS-SAR and VIS-IR.}
\label{tab.3}
\centering
\begin{minipage}{\textwidth}
\centering
\resizebox{\textwidth}{!}{%
\begin{tabular}{|c|c|c|c|c|c|c|c|c|}
\hline
Dataset & Threshold & RIFT & SDNet & ReDFeat & LoFTR & XoFTR & JamMa & RegNC-Mamba3\\
\hline
\multirow{3}{*}{VIS-SAR}
& @ 1px & 0.5 & 1.9 & \underline{6.1} & 2.8 & 0 & 0.9 & \textbf{17.7}\\
& @ 3px & 7.3 & 28.3 & 54.0 & \underline{51.9} & 29.2 & 29.5 & \textbf{84.2}\\
& @ 5px & 14.4 & 55.0 & \underline{75.7} & 74.8 & 60.1 & 60.4 & \textbf{97.6}\\
\hline
\multirow{3}{*}{VIS-IR}
& @ 1px & 11.3 & 22.5 & 52.6 & 64.6 & \underline{65.7} & 53.1 & \textbf{68.7}\\
& @ 3px & 58.7 & 76.5 & 92.5 & 91.3 & 94.5 & \underline{96.2} & \textbf{96.8}\\
& @ 5px & 71.8 & 91.0 & 96.2 & 94.6 & 96.9 & \underline{98.1} & \textbf{98.7}\\
\hline
\end{tabular}%
}
\end{minipage}
\end{table}

RegNC-Mamba3 obtains the lowest aRMSE and the highest aNCM and SMR on both datasets. On VIS-SAR, aRMSE is 1.901 and SMR is 97.6\%; the next-best aRMSE is 2.400 from ReDFeat, and the next-best SMR is 75.7\%. On VIS-IR, aRMSE decreases to 0.911 and SMR reaches 98.7\%. Table~\ref{tab.3} further shows the proportion of registered pairs below 1-, 3-, and 5-pixel RMSE thresholds. RegNC-Mamba3 has the highest proportion at every threshold on both datasets, including 17.7\% and 68.7\% below the 1-pixel threshold for VIS-SAR and VIS-IR.

\begin{table}[pos=H]
\caption{Registration results of semi-dense methods under different match budgets on VIS-SAR and VIS-IR.}
\label{tab.4}
\centering
\begin{minipage}{\textwidth}
\centering
\small
\renewcommand{\arraystretch}{0.92}
\begin{tabular}{|c|c|c|c|c|c|c|}
\hline
Dataset & Top-K & Metric & LoFTR & XoFTR & JamMa & RegNC-Mamba3\\
\hline
\multirow{9}{*}{VIS-SAR}
& \multirow{3}{*}{K=50}
& SMR(\%) & \underline{64.2} & 15.3 & 34.7 & \textbf{93.4}\\
& & aRMSE & \underline{2.707} & 3.603 & 3.420 & \textbf{2.208}\\
& & NCM & \textbf{31.1} & \underline{20.7} & 19.4 & 20.2\\
\cline{2-7}
& \multirow{3}{*}{K=500}
& SMR(\%) & \underline{74.1} & 53.3 & 60.8 & \textbf{97.6}\\
& & aRMSE & \underline{2.557} & 3.314 & 3.049 & \textbf{1.926}\\
& & NCM & \underline{180.9} & 148.8 & 91.1 & \textbf{185.3}\\
\cline{2-7}
& \multirow{3}{*}{K=2000}
& SMR(\%) & \underline{74.8} & 60.1 & 60.4 & \textbf{97.6}\\
& & aRMSE & \underline{2.584} & 3.075 & 3.033 & \textbf{1.901}\\
& & NCM & 281.4 & \underline{292.4} & 93.0 & \textbf{724.3}\\
\hline
\multirow{9}{*}{VIS-IR}
& \multirow{3}{*}{K=50}
& SMR(\%) & \underline{93.6} & 61.0 & 67.3 & \textbf{97.7}\\
& & aRMSE & \textbf{0.969} & 2.077 & 2.441 & \underline{1.126}\\
& & NCM & \textbf{46.0} & 32.0 & 24.4 & \underline{44.7}\\
\cline{2-7}
& \multirow{3}{*}{K=500}
& SMR(\%) & 94.9 & 92.7 & \underline{97.7} & \textbf{98.5}\\
& & aRMSE & \textbf{0.936} & 1.302 & 1.383 & \underline{0.972}\\
& & NCM & \underline{403.3} & 311.6 & 218.4 & \textbf{432.2}\\
\cline{2-7}
& \multirow{3}{*}{K=2000}
& SMR(\%) & 94.6 & 96.9 & \underline{98.1} & \textbf{98.7}\\
& & aRMSE & 0.983 & \underline{0.942} & 1.119 & \textbf{0.911}\\
& & NCM & 1148.3 & \underline{1222.8} & 732.4 & \textbf{1625.0}\\
\hline
\end{tabular}
\end{minipage}
\end{table}

NCM is not directly comparable between sparse and semi-dense methods because they produce different numbers of candidate correspondences. We therefore compare the semi-dense methods under match budgets $K\in\{50,500,2000\}$ in Table~\ref{tab.4}. Each method retains at most its top-$K$ highest-confidence correspondences, and all available matches are retained when fewer than $K$ are produced. RegNC-Mamba3 has the highest SMR at every budget on both datasets. At $K=50$, its SAR SMR is 93.4\% and its IR SMR is 97.7\%; XoFTR falls to 15.3\% and 61.0\%, while JamMa falls to 34.7\% and 67.3\%. The performance reduction is smaller for RegNC-Mamba3, indicating that its confidence ranking remains discriminative when the match budget is tight. On VIS-SAR, LoFTR, XoFTR, and JamMa often return fewer than 2,000 matches, whereas RegNC-Mamba3 produces more than 2,000 matches before truncation, which reflects the effectiveness of our coarse matching candidates.

\subsubsection{Performance on Four Image Pairs}

\begin{table}[pos=H]
\caption{RMSE and NCM of RegNC-Mamba3 and six comparison methods on four image pairs.}
\label{tab.5}
\centering
\begin{minipage}{\textwidth}
\centering
\begin{tabular}{|c|c|c|c|c|c|c|c|c|}
\hline
    Pairs & Metrics & RIFT &SDNet & ReDFeat & LoFTR & XoFTR & JamMa & RegNC-Mamba3\\
\hline
\multirow{2}{*}{pair 1}
& RMSE$\downarrow$ & 10 & 10 & 10 &\underline{1.88} & 10 & 10  & \textbf{1.26} \\

& NCM$\uparrow$ & 8 & 5 & 21 & \underline{214} & 35 & 34 & \textbf{360}\\
\hline
\multirow{2}{*}{pair 2}
& RMSE$\downarrow$ & 10 & 2.21 & \underline{1.57} & 2.11 & 2.13 &2.47 & \textbf{1.29} \\

& NCM$\uparrow$ & 35 & 191 & 282  & \underline{331} & 260 & 100 & \textbf{392}\\
\hline

\multirow{2}{*}{pair 3}
& RMSE$\downarrow$ & 4.22 & 3.04 & 1.76 & \underline{1.69} & 1.73 &1.79& \textbf{1.40} \\

& NCM$\uparrow$ & 107 & 409 & 461 & 871 & \underline{1363} & 1019 & \textbf{1503}\\
\hline

\multirow{2}{*}{pair 4}
& RMSE$\downarrow$ & 5.11 & 2.30 & \underline{2.06} & 2.16 & 2.20 & 2.08& \textbf{1.39} \\

& NCM$\uparrow$ & 731 & 417 & 960 & 1024 & \underline{1546} & 1284 & \textbf{1603}\\
\hline

\end{tabular}
\end{minipage}
\end{table}

\begin{figure}[H]
\centering
\includegraphics[width=\textwidth,height=0.70\textheight,keepaspectratio]{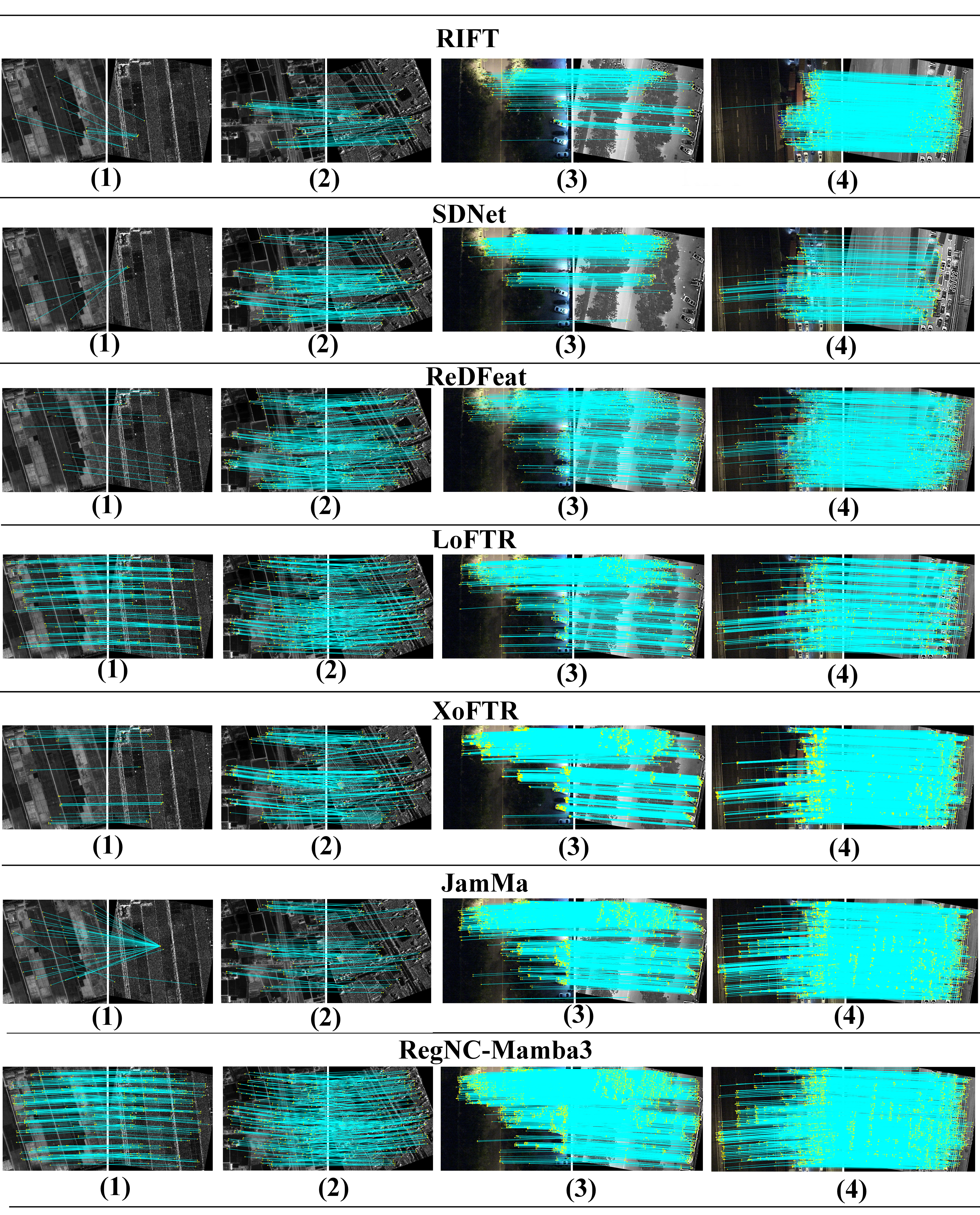}
\caption{Visualization of matching performance. Only the inlier matches after MAGSAC are shown.}
\label{fig.5}
\end{figure}

\begin{figure}[H]
\centering
\includegraphics[width=1\textwidth,keepaspectratio]{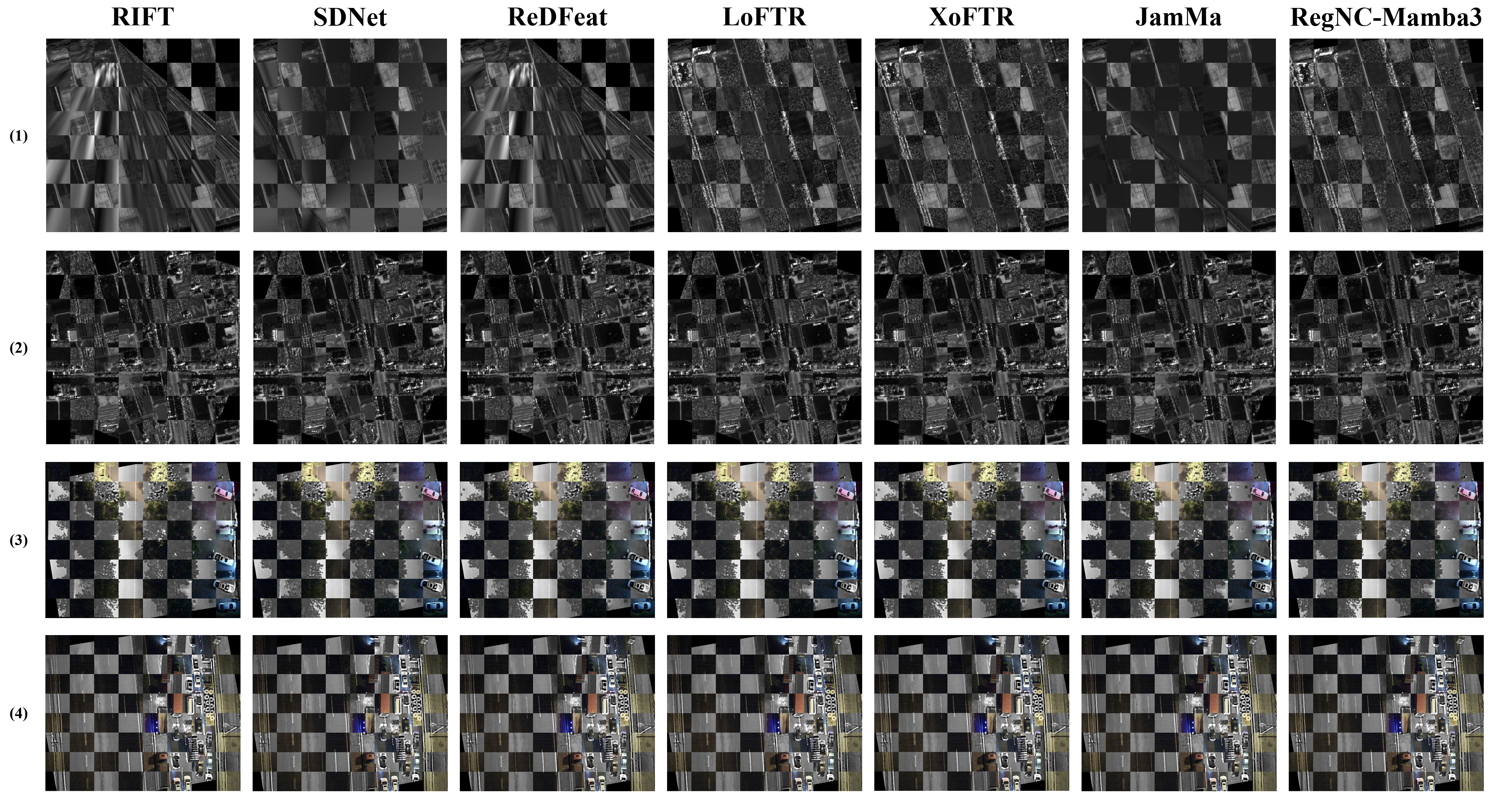}
\vspace{-25pt}
\caption{Checkerboard images of Pair 1 to Pair 4 registration results.}
\label{fig.6}
\end{figure}

\begin{figure}[H]
\centering
\includegraphics[width=1\textwidth,keepaspectratio]{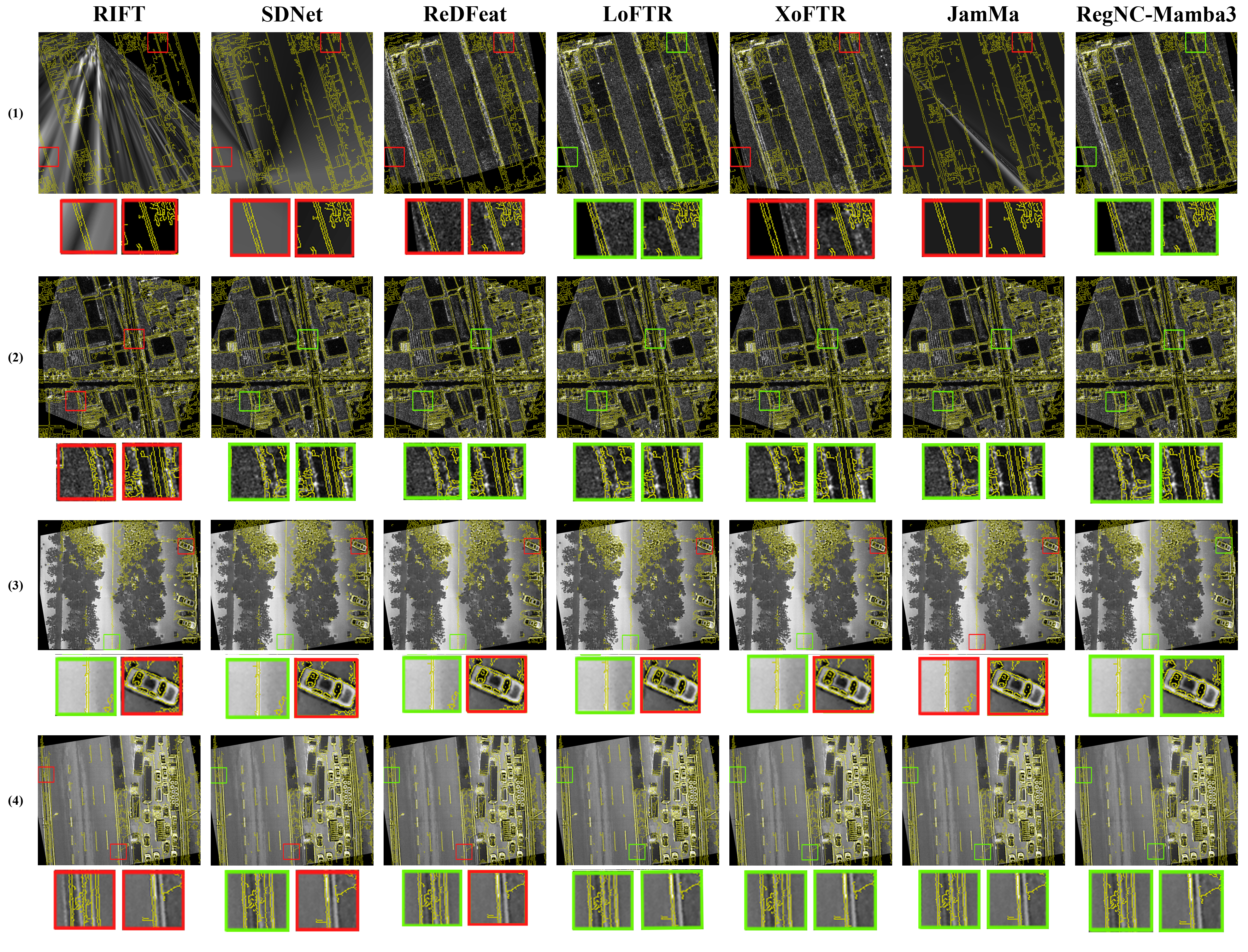}
\vspace{-25pt}
\caption{Edge images of Pair 1 to Pair 4 registration results. Red and green rectangles circle the areas not aligned or aligned.}
\label{fig.7}
\end{figure}

Four representative pairs are used for qualitative and pair-level analysis. Pairs~1 and~2 are selected from OSDataset (VIS-SAR), and Pairs~3 and~4 from DroneVehicle (VIS-IR). A geometric transformation is applied before registration. Table~\ref{tab.5} reports NCM and RMSE for each pair, and an RMSE of 10 denotes a failed registration. Figure~\ref{fig.5} shows the retained matches after MAGSAC. Figures~\ref{fig.6} and~\ref{fig.7} present checkerboard and edge overlays with red and green rectangles marking misaligned and aligned regions.

Pair~1 is a difficult VIS-SAR scene with few local details and strong speckle noise. RIFT and SDNet operate on local patches that contain few reliable edges, so neither handcrafted nor CNN-based descriptors are sufficiently discriminative. ReDFeat avoids patch-level keypoint detection but still lacks global structural aggregation and cross-modal interaction. XoFTR and JamMa fail to produce enough reliable coarse correspondences, and the subsequent local refinement cannot recover from this failure. In Table~\ref{tab.5}, only LoFTR and RegNC-Mamba3 register Pair~1 successfully; their RMSE values are 1.88 and 1.26.

Pair~2 contains more detail and includes a $20^{\circ}$ rotation. RIFT underperforms because its handcrafted representation is less robust to rotation in this scene, whereas the other methods register the pair successfully. RegNC-Mamba3 obtains the lowest RMSE and highest NCM on both VIS-SAR pairs. Its features are produced by NC-SSP over all scales, and the Log-Gabor modulation emphasizes structural features, which help retain correspondences under speckle and rotation.

Pairs~3 and~4 are night-time VIS-IR examples. The visible images have weaker edge information, which reduces both the number and the spatial uniformity of keypoints detected by RIFT and SDNet. Infrared images contain less speckle than SAR, so ReDFeat and all semi-dense methods register Pairs~3 and~4 successfully. RegNC-Mamba3 again obtains the lowest RMSE and highest NCM on both pairs, with RMSE values of 1.40 and 1.39. Across the four examples, its advantage is most pronounced on the noisy VIS-SAR pair, demonstrating that RegNC-Mamba3 provides stronger global aggregation and noise suppression.

\subsection{Complexity and Efficiency Analysis}
For detect-then-describe methods, we record keypoint detection, description, and feature matching time; for end-to-end methods, we record total network runtime. MAGSAC is excluded from all timing measurements. Measurements use an Intel i9-13900K CPU and an NVIDIA GeForce RTX 4090 with 24 GB of memory. RIFT runs entirely on the CPU in MATLAB and has no GPU implementation. SDNet performs handcrafted keypoint detection on the CPU and CNN descriptor extraction on the GPU. ReDFeat, LoFTR, XoFTR, JamMa, and RegNC-Mamba3 run end-to-end on the GPU. Table~\ref{tab.6} reports FLOPs, runtime, GPU memory, and aRMSE for inputs of $512\times512$ (VIS-SAR) and $640\times512$ (VIS-IR).

\begin{table}[pos=H]
\caption{Registration accuracy and computational cost of the compared methods.}
\label{tab.6}
\centering
\begin{minipage}{\textwidth}
\centering
\resizebox{\textwidth}{!}{%
\begin{tabular}{|c|c|c|c|c|c|c|c|c|}
\hline
\multirow{2}{*}{\diagbox{Methods}{Datasets}} & \multicolumn{4}{c|}{VIS-SAR} & \multicolumn{4}{c|}{VIS-IR} \\
\cline{2-9}
& aRMSE$\downarrow$ & FLOPs (G)$\downarrow$ &time(ms)&memory(MB) & aRMSE$\downarrow$ &FLOPs (G)$\downarrow$ &time(ms)&memory(MB)  \\
\hline
RIFT* & 3.104&-&5683&-&2.009&-&5730&-\\
SDNet* & 2.984&481&411&174&1.822&630&522&174\\
\hline
ReDFeat &\underline{2.400}&874.2&53.40&\underline{713}&1.168&1092.3&82.32&\underline{890}\\
LoFTR & 2.584 & 659.7 & \underline{37.87} & 906 & 0.983 & 916.6 & \underline{62.85} & 1104\\
XoFTR & 3.075 & \underline{624.4} & 60.09 & 1967 & \underline{0.942} & 806.6 & 82.30 & 2626\\
JamMa & 3.033 & \textbf{173.1} & \textbf{31.93} & \textbf{687} & 1.119 & \textbf{215.1} & \textbf{34.91} & \textbf{831}\\
RegNC-Mamba3 &\textbf{1.901} &627.1 & 93.54 & 765 &\textbf{ 0.911} & 813.0 & 129.45 & 906\\
\hline
\end{tabular}%
}
\end{minipage}
\end{table}

RIFT and SDNet have much longer runtimes because original phase-congruency and FAST detection are not GPU-accelerated. For SDNet, the reported FLOPs and GPU memory cover only descriptor extraction from 5,000 patches; CPU keypoint detection is excluded. The asterisk in Table~\ref{tab.6} marks methods that are not fully GPU-executed. Their FLOPs and memory values are listed for reference only and are not included in the comparison.

Among the GPU methods, ReDFeat has the highest FLOP count, whereas JamMa has the lowest FLOP count, runtime, and memory cost, but its aRMSE remains 3.033 on VIS-SAR and 1.119 on VIS-IR. RegNC-Mamba3 uses fewer FLOPs than LoFTR and slightly more than XoFTR on both datasets, while its measured runtime is 93.54 ms and 129.45 ms. This gap shows that FLOPs alone do not determine the inference speed. Multi-scale NC-SSP involves matrix operations at different resolutions; kernel-launch overhead and memory-bandwidth limitations of NC-SSP may contribute to the increased inference latency. Relative to the four-directional vanilla Mamba-3 baseline, NC-SSP remains much faster and uses less GPU memory. RegNC-Mamba3 incurs longer latency than other semi-dense methods in exchange for lower registration error and a higher successful matching rate.

\section{Conclusion}

This paper presented RegNC-Mamba3 for multi-modal image registration. The central component is NC-SSP, which converts the vanilla state space principles into a scan-path-invariant form to achieve non-causality. Relative-weight normalization removes long-range decay, and the hidden state with aggregation over all circular scan orders allows every token to receive global context without relying on a fixed scan direction. We further introduced structural feature awareness by constructing a complex-valued hidden state with learnable multi-orientation Log-Gabor filters. In RegNC-Mamba3, the NC-SSP blocks are used for cross-modal interaction at each scale and for multi-scale feature fusion within a coarse-to-fine matching network.

Experiments on VIS-SAR and VIS-IR show that RegNC-Mamba3 achieves the lowest aRMSE and the highest aNCM and SMR among the compared methods. It obtains aRMSE values of 1.901 and 0.911 and SMR values of 97.6\% and 98.7\% on the two datasets. The ablation study shows that NC-SSP benefits from combining path-invariant aggregation with structural feature awareness. NC-SSP also reduces runtime and GPU memory relative to the four-directional vanilla Mamba-3 baseline, although the complete network remains slower than the lightweight JamMa and has longer inference time than LoFTR and XoFTR. Reducing this overhead while retaining the registration gain is a direction for future work.

\section*{Acknowledgments}
This work was supported in part by the Natural Science Foundation of China under Grant 62172321. (Corresponding author: Yan Wu.)

\section*{Declaration of competing interests}
There are no competing financial interests or personal relationships that could affect the work presented in this paper.

\printcredits




\begin{thebibliography}{99}



\bibitem{ref1} C. Liu, X. Ma, X. Yang, et al., COMO: Cross-Mamba Interaction and Offset-Guided Fusion for Multimodal Object Detection, Inf. Fusion 125 (2026) 103414.

\bibitem{ref2} Z. Xiong, W. Li, X. Zhao, et al., PRF-Net: A Progressive Remote Sensing Image Registration and Fusion Network, IEEE Trans. Neural Netw. Learn. Syst. 36 (5) (2025) 9437--9450.

\bibitem{ref3} B. Li, X. Li, B. Wang, et al., MambaCFFuse: Infrared and Visible Image Fusion Based on VMamba with Complementary Feature Aggregation, Appl. Soft Comput. 183 (2025) 113629.

\bibitem{ref4} X. Xin, M. Li, Y. Wu, et al., Semi-Supervised Classification of Dual-Frequency PolSAR Image Using Joint Feature Learning and Cross Label-Information Network, IEEE Trans. Geosci. Remote Sens. 60 (2022) 1--16.


\bibitem{ref5} Z. Yang, Y. Wu, M. Li, et al., Unsupervised Difference-Guided Adaptive Fusion Network for Change Detection in PolSAR Images, IEEE Trans. Geosci. Remote Sens. 62 (2024) 1--16.

\bibitem{ref6} W. Cheng, Y. Feng, Y. Sun, et al., Heterogeneous Remote Sensing Image Change Detection Network Based on Multi-scale Feature Modal Transformation, Appl. Soft Comput. 170 (2025) 112725.

\bibitem{ref7} Z. Fan, Y. Pi, M. Wang, et al., GLS--MIFT: A modality invariant feature transform with global-to-local searching, Inf. Fusion 105 (2024) 102252.

\bibitem{ref8} Z. Hou, Y. Liu, L. Zhang, POS-GIFT: A geometric and intensity-invariant feature transformation for multimodal images, Inf. Fusion 102 (2023) 102027.

\bibitem{ref9} Y. Liao, P. Tao, Q. Chen, et al., Highly adaptive multi-modal image matching based on tuning-free filtering and enhanced sketch features, Inf. Fusion 112 (2024) 102599.

\bibitem{ref10} J. Li, Q. Hu, M. Ai, RIFT: Multi-Modal Image Matching Based on Radiation-Variation Insensitive Feature Transform, IEEE Trans. Image Process. 29 (2019) 3296--3310.

\bibitem{ref11} Y. Yao, Y. Zhang, Y. Wan, et al., Multi-Modal Remote Sensing Image Matching Considering Co-Occurrence Filter, IEEE Trans. Image Process. 31 (2022) 2584--2597.

\bibitem{ref12} D. Quan, S. Wang, Y. Li, et al., Multi-Relation Attention Network for Image Patch Matching, IEEE Trans. Image Process. 30 (2021) 7127--7142.




\bibitem{ref13} X. Zhang, Y. Sun, Y. Han, et al., SSML-QNet: Scale-Separative Metric Learning Quadruplet Network for Multi-modal Image Patch Matching, in: Proceedings of the Thirty-Second International Joint Conference on Artificial Intelligence, 2023, pp. 4593--4601.

\bibitem{ref14} A. Barroso-Laguna, K. Mikolajczyk, Key.Net: Keypoint Detection by Handcrafted and Learned CNN Filters Revisited, IEEE Trans. Pattern Anal. Mach. Intell. 45 (1) (2023) 698--711.


\bibitem{ref15} S. Zhang, Z. Fu, J. Liu, et al., Multilevel Attention Siamese Network for Keypoint Detection in Optical and SAR Images, IEEE Trans. Geosci. Remote Sens. 61 (2023) 1--17.



\bibitem{ref16} J. Revaud, C. De Souza, M. Humenberger, et al., R2D2: Reliable and Repeatable Detector and Descriptor, Adv. Neural Inf. Process. Syst. 32 (2019).

\bibitem{ref17} Y. Deng, J. Ma, ReDFeat: Recoupling Detection and Description for Multimodal Feature Learning, IEEE Trans. Image Process. 32 (2023) 591--602.









\bibitem{ref18} Z. Li, J. Jiang, J. Ma, Seed to prune: A seeded graph neural network for two-view correspondence learning, IEEE Trans. Neural Netw. Learn. Syst. 36 (7) (2025) 12199--12213.

\bibitem{ref19} J. Sun, Z. Shen, Y. Wang, et al., LoFTR: Detector-Free Local Feature Matching with Transformers, in: Proceedings of the IEEE/CVF Conference on Computer Vision and Pattern Recognition, 2021, pp. 8922--8931.


\bibitem{ref20} H. Chen, Z. Luo, L. Zhou, et al., ASpanFormer: Detector-Free Image Matching with Adaptive Span Transformer, in: European Conference on Computer Vision, 2022, pp. 20--36.


\bibitem{ref21} Ö. Tuzcuoglu, A. Koksal, B. Sofu, et al., XoFTR: Cross-modal Feature Matching Transformer, in: Proceedings of the IEEE/CVF Conference on Computer Vision and Pattern Recognition, 2024, pp. 4275--4286.


\bibitem{ref22} Y. Zhang, S. Yi, H. Wu, et al., CEFG-Net: Context-enhanced features guided network for infrared and visible UAV image stitching, Appl. Soft Comput. 195 (2026) 115045.


\bibitem{ref23} J. Edstedt, Q. Sun, G. Bökman, et al., RoMa: Robust dense feature matching, in: Proceedings of the IEEE/CVF Conference on Computer Vision and Pattern Recognition, 2024, pp. 19790--19800.


\bibitem{ref24} M.A. Fischler, R.C. Bolles, Random Sample Consensus: A Paradigm for Model Fitting With Applications to Image Analysis and Automated Cartography, Commun. ACM 24 (6) (1981) 381--395.

\bibitem{ref25} D. Barath, J. Matas, J. Noskova, MAGSAC: Marginalizing Sample Consensus, in: Proceedings of the IEEE/CVF Conference on Computer Vision and Pattern Recognition, 2019, pp. 10197--10205.

\bibitem{ref26} E. Shojaedini, M. Majd, R. Safabakhsh, Novel adaptive genetic algorithm sample consensus, Appl. Soft Comput. 77 (2019) 635--642.


\bibitem{ref27} A. Gu, T. Dao, Mamba: Linear-Time Sequence Modeling with Selective State Spaces, in: First Conference on Language Modeling, 2024.

\bibitem{ref28} Y. Liu, Y. Tian, Y. Zhao, et al., VMamba: Visual State Space Model, Adv. Neural Inf. Process. Syst. 37 (2024) 103031--103063.

\bibitem{ref29} T. Huang, X. Pei, S. You, et al., LocalMamba: Visual State Space Model with Windowed Selective Scan, in: European Conference on Computer Vision, 2024, pp. 12--22.

\bibitem{ref30} Y. Shi, M. Dong, C. Xu, Multi-Scale VMamba: Hierarchy in Hierarchy Visual State Space Model, Adv. Neural Inf. Process. Syst. 37 (2024) 25687--25708.

\bibitem{ref31} D. Han, Z. Wang, Z. Xia, et al., Demystify Mamba in Vision: A Linear Attention Perspective, Adv. Neural Inf. Process. Syst. 37 (2024) 127181--127203.

\bibitem{ref32} T. Dao, A. Gu, Transformers are SSMs: Generalized Models and Efficient Algorithms Through Structured State Space Duality, Proc. Mach. Learn. Res. 235 (2024) 10041--10071.

\bibitem{ref33} Y. Shi, M. Dong, M. Li, et al., VSSD: Vision Mamba with Non-Causal State Space Duality, in: Proceedings of the IEEE/CVF International Conference on Computer Vision, 2025, pp. 10819--10829.

\bibitem{ref34} A. Lahoti, K.Y. Li, B. Chen, et al., Mamba-3: Improved Sequence Modeling using State Space Principles, in: International Conference on Learning Representations, 2026.

\bibitem{ref35} A. Ramachandran, et al., Vision Non-Causal Trapezoidal Mamba: Eliminating Directional Scanning in Vision SSMs with Second-Order Dynamics, arXiv preprint arXiv:2607.03589v2, 2026.

\bibitem{ref36} E. Rosten, T. Drummond, Machine Learning for High-Speed Corner Detection, in: European Conference on Computer Vision, 2006, pp. 430--443.

\bibitem{ref37} D. Quan, H. Wei, S. Wang, et al., Self-distillation feature learning network for optical and SAR image registration, IEEE Trans. Geosci. Remote Sens. 60 (2022) 1--18.

\bibitem{ref38} Q. Zhu, et al., Rethinking Scanning Strategies with Vision Mamba in Semantic Segmentation of Remote Sensing Imagery: An Experimental Study, IEEE J. Sel. Top. Appl. Earth Obs. Remote Sens. 17 (2024) 18223--18234.

\bibitem{ref39} X. Lu, S. Du, JamMa: Ultra-lightweight Local Feature Matching with Joint Mamba, in: Proceedings of the IEEE/CVF Conference on Computer Vision and Pattern Recognition, 2025, pp. 14934--14943.


\bibitem{ref40} J. Arrospide, L. Salgado, Log-Gabor Filters for Image-Based Vehicle Verification, IEEE Trans. Image Process. 22 (6) (2013) 2286--2295.

\bibitem{ref41} Y. Xiang, R. Tao, F. Wang, et al., Automatic Registration of Optical and SAR Images Via Improved Phase Congruency Model, IEEE J. Sel. Top. Appl. Earth Obs. Remote Sens. 13 (2020) 5847--5861.

\bibitem{ref42} Y. Sun, B. Cao, P. Zhu, et al., Drone-Based RGB-Infrared Cross-Modality Vehicle Detection via Uncertainty-Aware Learning, IEEE Trans. Circuits Syst. Video Technol. 32 (10) (2022) 6700--6713.


\end{thebibliography}
\end{document}